\documentclass[10pt, conference, compsocconf]{IEEEtran}

\usepackage{cite}
\usepackage{mathtools}
\ifCLASSINFOpdf
  \usepackage[pdftex]{graphicx}
\else
\fi
\begin{document}

\title{A Classification Refinement Strategy for Semantic Segmentation}

\author{\IEEEauthorblockN{James W. Davis, Christopher Menart, Muhammad Akbar}
\IEEEauthorblockA{Dept. of Comp. Sci. \& Eng., Ohio State University\\
Columbus, OH USA\\
Email: \{davis.1719, menart.1, akbar.24\}@osu.edu}
\and
\IEEEauthorblockN{Roman Ilin}
\IEEEauthorblockA{RYAP, Air Force Research Laboratory\\
Wright Patterson AFB, OH USA\\
Email: rilin325@gmail.com}
}

\maketitle


\begin{abstract}
Based on the observation that semantic segmentation  errors are partially predictable, we propose a compact formulation using confusion statistics of the trained classifier to refine (re-estimate) the initial pixel label hypotheses.  The proposed strategy is contingent upon computing the classifier confusion probabilities for a given dataset and estimating a relevant prior on the object classes present in the image to be classified. We provide a procedure to robustly estimate the confusion probabilities and explore multiple prior definitions. Experiments are shown comparing performances on multiple challenging datasets using different priors to improve a state-of-the-art semantic segmentation classifier. This study demonstrates the potential to significantly improve semantic labeling and motivates future work for reliable label prior estimation from images.  
\end{abstract}


\IEEEpeerreviewmaketitle

\section{Introduction}

Semantic segmentation is a challenging computer vision problem wherein a class label is assigned to each pixel in an image. This provides much richer scene information than traditional image classification, and therefore is inherently a more difficult task. While image classification requires the detection of the primary object, semantic segmentation requires precise, localized pixel-wise detections. Annotating training images for semantic segmentation in a supervised manner is more expensive and time-consuming, hence most public datasets are relatively small in comparison to those for image classification (e.g., ImageNet \cite{Deng1}).


Current semantic segmentation methods show promising results on multiple datasets, however systematic errors in their output can be readily found. For example, RefineNet \cite{Lin3} mis-classifies {\tt ground} as {\tt sidewalk} so often in the PASCAL-Context dataset \cite{Mottaghi14} that pixels classified as {\tt sidewalk} are 60\% likely to actually be {\tt ground}. In this  example, pixel accuracy would be improved by simply re-labeling all {\tt sidewalk} pixels as {\tt ground}. However, it is more preferable to use this confusion knowledge to better reason about certain situations in a probabilistic manner.

\begin{figure}[t]
\begin{center}
\begin{tabular}{c|c}
\includegraphics[width=1.5in]{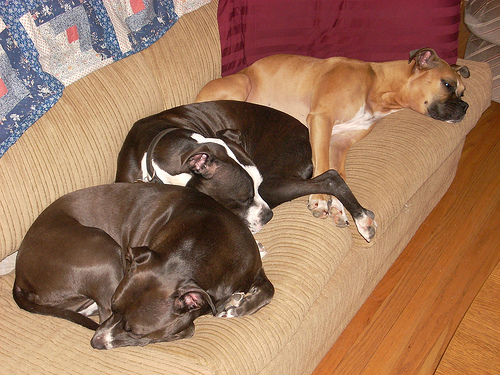}&
\includegraphics[width=1.5in]
{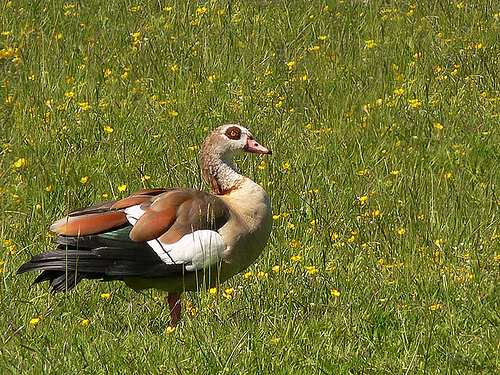}\\
\includegraphics[width=1.5in]
{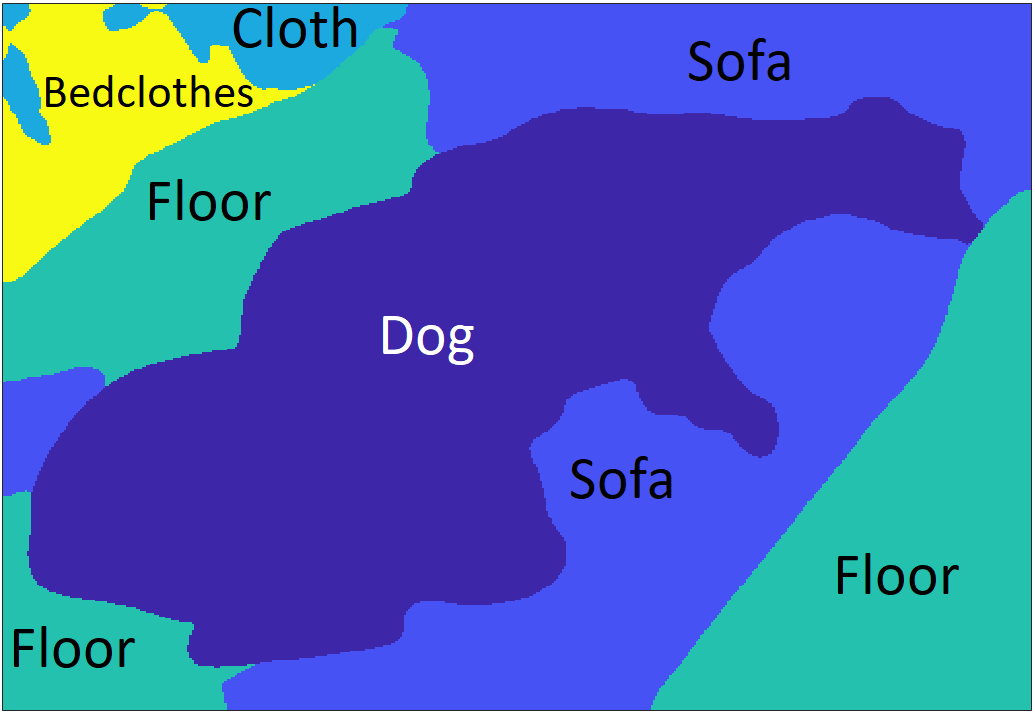}&
\includegraphics[width=1.5in]
{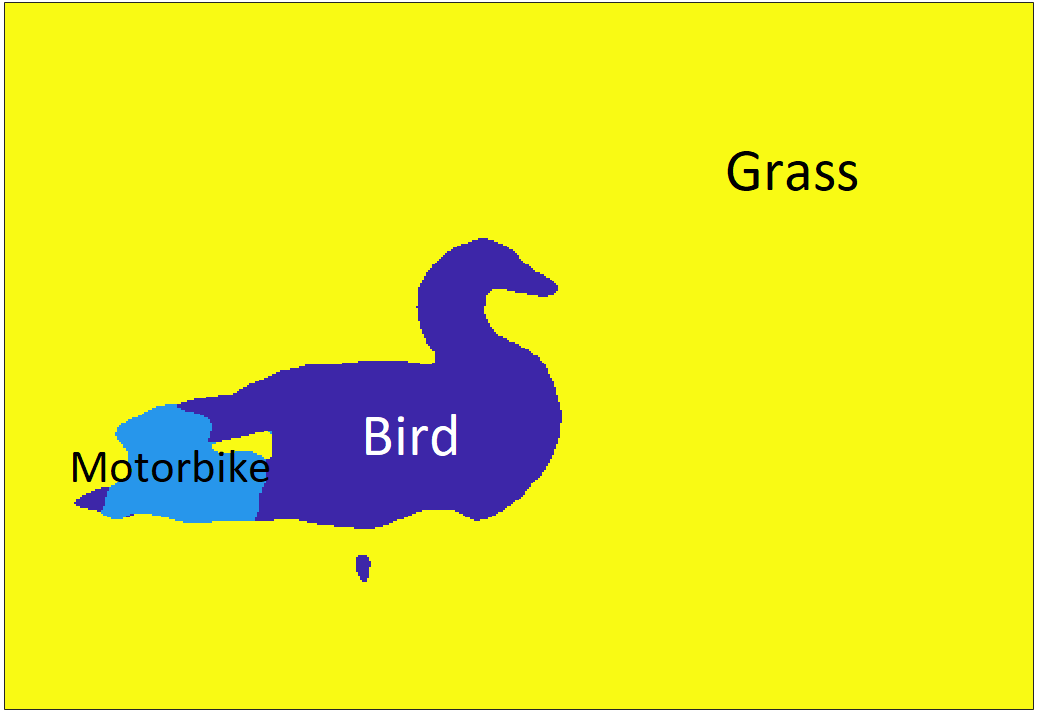}\\
\end{tabular}
\end{center}
\caption{Example contextual errors of RefineNet on PASCAL-Context. {\em In-Context}: The sofa back in the left image is classified as {\tt floor}, which is incorrect even though the label correctly appears elsewhere in the image. {\em Out-of-Context}: A portion of the right image is classified as {\tt motorbike}, which is not a valid label anywhere in the image.}
\label{fig:mistakes}
\end{figure}

We categorize labeling errors into one of two types of mistakes. An ``{\em in-context}'' labeling error is when a pixel is assigned an incorrect label from the set of actual labels for the given image (e.g., a foreground object label is assigned to an incorrect location). An ``{\em out-of-context}'' labeling error is an invalid pixel label assignment that is outside of the actual label set for the image (e.g., an indoor object label is assigned to a pixel in an image of an outdoor scene). Actual examples of these errors are shown in Fig.~\ref{fig:mistakes}.

Given that a classifier will have similar performance characteristics on related images, analysis of the classification errors and label confusions across a dataset should support a secondary (post-processing) refinement stage to re-estimate the output label confidences to better classify the pixels, reducing both in-context and out-of-context errors. The proposed approach is based on these  underlying motivations.

Our method is derived from a direct marginalization of $p(l | d)$, the probability of label $l$ given input data $d$, resulting in a decomposition of the formulation into classifier output label probabilities and learned classifier/truth confusions. Applied together, the framework treats the confidence levels of the original classifier output as partial evidence, and incorporates the confusion information to determine the final probability of witnessing each object label at each location in an image. In a sense, this confusion approach can be considered a form of context-aware re-estimation. We present a robust method to compute the confusion probabilities and also outline various label priors used to bias the refinement. Upper bound performances of the framework with various priors are reported  for three challenging datasets to justify the approach.

The rest of this paper is organized as follows. Section
\ref{Sect:RelatedWork} describes related work in semantic segmentation. Section \ref{Sect:Approach} describes our framework for classifier refinement using label confusion probabilities and priors. Section \ref{Sect:Experiments} presents experimental results  showing the performance improvements of the approach on multiple datasets, followed by a conclusion in Sect.~\ref{Sect:Conclusion}


\section{Related Work}
\label{Sect:RelatedWork}

Convolutional Neural Networks (CNNs) such as \cite{Long1, Chen1, Lin1, Lin3} have achieved unprecedented results in semantic segmentation. However, they also introduce a tension between precise localization and inclusion of broader image context \cite{Garcia-Garcia17}. Many recent innovations in network architecture were motivated by this issue, including dilated convolution \cite{Chen1}, multi-scale prediction \cite{Raj1}, zoom-out features \cite{Mostajabi15}, recursive convolutional networks \cite{Visin1}, and convolutional upsampling \cite{Lin3,Ronneberger1}.


Context is exploited using additional techniques, such as K-nearest-neighbors methods \cite{Tighe1}, conditional random fields (CRFs) \cite{Shotton1,Krahenbuhl1}, and precision matrices \cite{Souly16}. These approaches remain relevant in part due to their flexibility. Research has continued to investigate estimating and employing co-occurrence statistics for CRFs \cite{Zhang1}, and some recent methods have attempted to integrate CRFs directly into neural models \cite{Zheng15,Arnab16}.
Our proposed method provides a Bayesian framework to incorporate context formed by confusion probabilities and label priors to influence pixel-level decisions.

The most related work to the proposed framework is the approach of LabelBank  \cite{Hu1}. Both methods process secondary information to adjust the confidence values produced by a base classifier. LabelBank  uses a CNN architecture to estimate which classes are present in the image, then essentially filters out labels in the output of the base semantic segmentation classifier that do not belong to this list. The method therefore focuses on the removal of out-of-context errors. Our strategy is more flexible, as we include additional confusion probabilities and relax the requirement of binary estimation of the classes, which together allow the method to be more general to remove both in-context and out-of-context errors. Furthermore, our approach is supported within a Bayesian framework, rather than applying an ad hoc masking technique. We will show in the experiments that the upper bound performance capability of our approach is above the LabelBank optimum.

As we describe our procedure to refine the classifier output label probabilities, one may be tempted to view this as a form of boosting. Similar to boosting, our approach is a staged process, first running the base classifier then refining the output hypotheses (two stages). However, boosting trains a series of multiple (weak) classifiers using re-weighted examples based on the errors from the previous stage(s).

\section{Approach}
\label{Sect:Approach}

Let $C$ be a given semantic segmentation classifier that generates pixel-wise label hypotheses for images. The output of $C$ at each pixel site provides a confidence/probability for every possible class (typically attained using softmax). For image data $d_i$ at each pixel site $i$ for each possible label $l \in \mathcal{L}$, we denote the classifier output probabilities  as
$P(C(d_i)= l  | d_i)$. Note that we differentiate this from the typical classifier output notation of $P(l | d_i)$. The final predictions are usually made by choosing the label that is assigned the highest confidence/probability at each site. 

We propose a Bayesian method to exploit the confusions exhibited by classifier $C$ to probabilistically refine and improve the estimate of the best label at each site. We begin by examining $P(l_{i}^{gt}  | d_i)$, the probability of site $i$ having its {\em ground truth} label ($l_{i}^{gt}$) given the image data $d_i$. Note that this probability is {\em not} the classifier output probability, and hence is currently independent of any particular classifier. 

A marginalization of this probability over the possible classifier outputs yields
{\setlength\arraycolsep{0.1em}
	\begin{eqnarray}
	P(l_{i}^{gt} | d_i)  & = & \sum_{c \in \mathcal{L}} P(l_{i}^{gt}, C(d_i) = c | d_i) \label{eqn:orig-form1} \\
	&  = & \sum_{c \in \mathcal{L}} P(l_{i}^{gt} | C(d_i) = c, d_i)  \cdot P(C(d_i) = c | d_i)  \label{eqn:orig-form2} \\
	&  = & \sum_{c \in \mathcal{L}} P(l_{i}^{gt}  | C(d_i) = c)  \cdot P(C(d_i) = c | d_i) \label{eqn:orig-form3}
	\end{eqnarray}
}

\noindent The first term in the two-part decomposition of Eqn.~\ref{eqn:orig-form3} represents the confusion relationship between the ground truth label and the classifier $C$ outputs. The second term contains the classifier's output probabilities, as initially described. 

Applying Bayes' Rule to the first term in Eqn.~\ref{eqn:orig-form3} yields
\begin{equation}
P(l_{i}^{gt} | C(d_i)= c)  =  \frac{P(C(d_i)= c | l_{i}^{gt}) \cdot P(l_{i}^{gt})}{P(C(d_i)= c)}
\label{eqn:bayes}
\end{equation}

\noindent where the denominator $P(C(d_i)= c)$ is computed from the integration of the numerator over all possible labels in the set $\mathcal{L}$. We relax these probabilities to be both site-independent and data-independent for ease in learning and for generalization. Hence the probability for any pixel (regardless of site location or image) having any label $l \in \mathcal{L}$ given the classifier's selected class label $c$  is 
\begin{equation}
P(l | C(d_i)= c)  =  \frac{P(C=c | l) \cdot P(l)}{P(C=c)} \label{eqn:bayes-site-indep}
\end{equation}

\noindent where $P(C=c | l)$ is the generalized confusion (or mismatch) probability between a given/known label and the classifier's output, and $P(l)$ is a prior on the label $l$. 

Merging Eqn.~\ref{eqn:bayes-site-indep} into Eqn.~\ref{eqn:orig-form3} yields the refinement, or re-estimation, formulation of the original site-{\em dependent} classifier output probabilities $P(C(d_i) = c | d_i), \forall c \in \mathcal{L}$ into more confusion-aware probabilities. The probability of {\em any} output label $l \in \mathcal{L}$ given the input data is now described by
\begin{equation}
	P(l | d_i)  = \sum_{c \in \mathcal{L}} \frac{P(C=c | l) \cdot P(l) \cdot P(C(d_i) = c | d_i) }{P(C=c)} \label{eqn:final-refine}
\end{equation}

\noindent Hence Eqn.~\ref{eqn:final-refine} can be used to determine the new refined probabilities for each possible label $l \in \mathcal{L}$ at pixel site $i$.

We can efficiently formulate this process employing a single transformation  matrix with all possible label combinations. An $|\mathcal{L}| \times |\mathcal{L}|$ refinement matrix $R$ is defined as $R(c_1, c_2) = P(c_1 | C=c_2)$, given by all $c_1, c_2 \in \mathcal{L}$ instantiations within Eqn.~\ref{eqn:bayes-site-indep}. For pixel site $i$ in the test image, we construct the vector of site-dependent base classifier output probabilities as $X_{i}(c) = P(C(d_i) = c | d_i), \forall c\in \mathcal{L}$. Following  Eqn.~\ref{eqn:final-refine}, $\forall l \in \mathcal{L}$, and employing matrix $R$, it is equivalent to refine the original classifier output probabilities with the linear transformation
\begin{equation}
\hat{X}_{i} = R \cdot X_{i}
\end{equation}

\noindent As usual, the selected class label for site $i$ will be the label/entry in $\hat{X}_i$ (instead of $X_{i}$) that is maximal. 

This refinement process is conducted for each site in the test image being classified. Since the matrix $R$ is computed only once (and fixed) for the entire image using site-independent data (with Eqn.~\ref{eqn:bayes-site-indep}), the re-estimation procedure is quite fast and computationally efficient. 

The refinement framework is straightforward to derive, but as we will show, can be used to significantly improve performance. We next present the means to robustly compute confusion probabilities and detail various forms of the prior expectations needed for this process.


\subsection{Confusion Probabilities}
\label{SubSect:confusion-probabilities}

The site-independent classifier/truth confusion (error) probabilities $P(C=c | l)$ required in Eqn.~\ref{eqn:final-refine} can be readily learned from data by collecting all sites (for all images employed) having a specific ground truth label  and determining the frequency of those sites the classifier assigned label $c \in \mathcal{L}$. To ensure all probabilities are defined, a minimum value of $10^{-4}$ is used for any confusion pairing (before the final normalization of counts into probabilities) if the number of sites is 0 for a particular classification output given the ground truth label. 

\begin{figure}[t]
\begin{center}
\begin{tabular}{cc}
\includegraphics[width=1.6in]{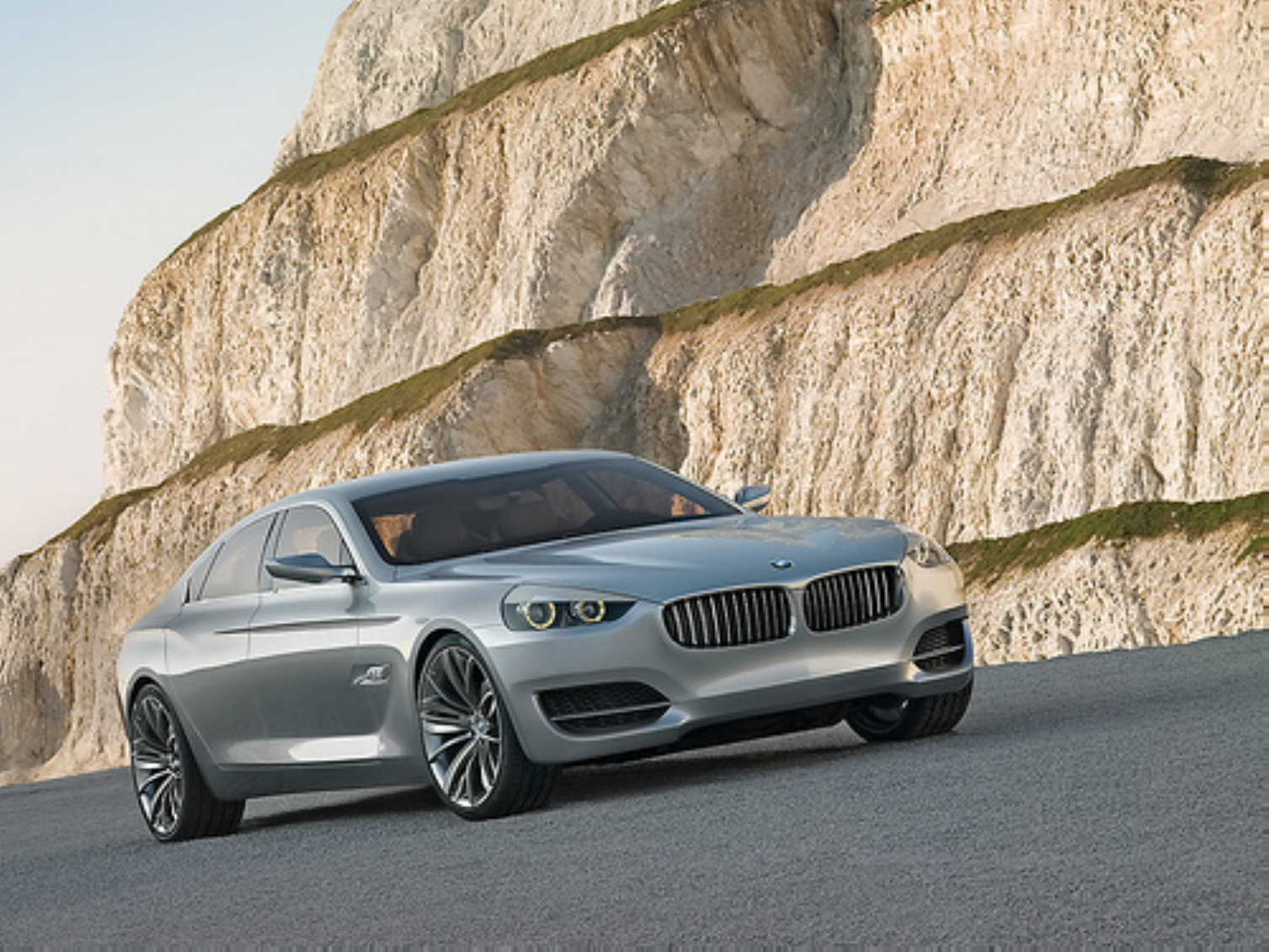}&
\includegraphics[width=1.6in]{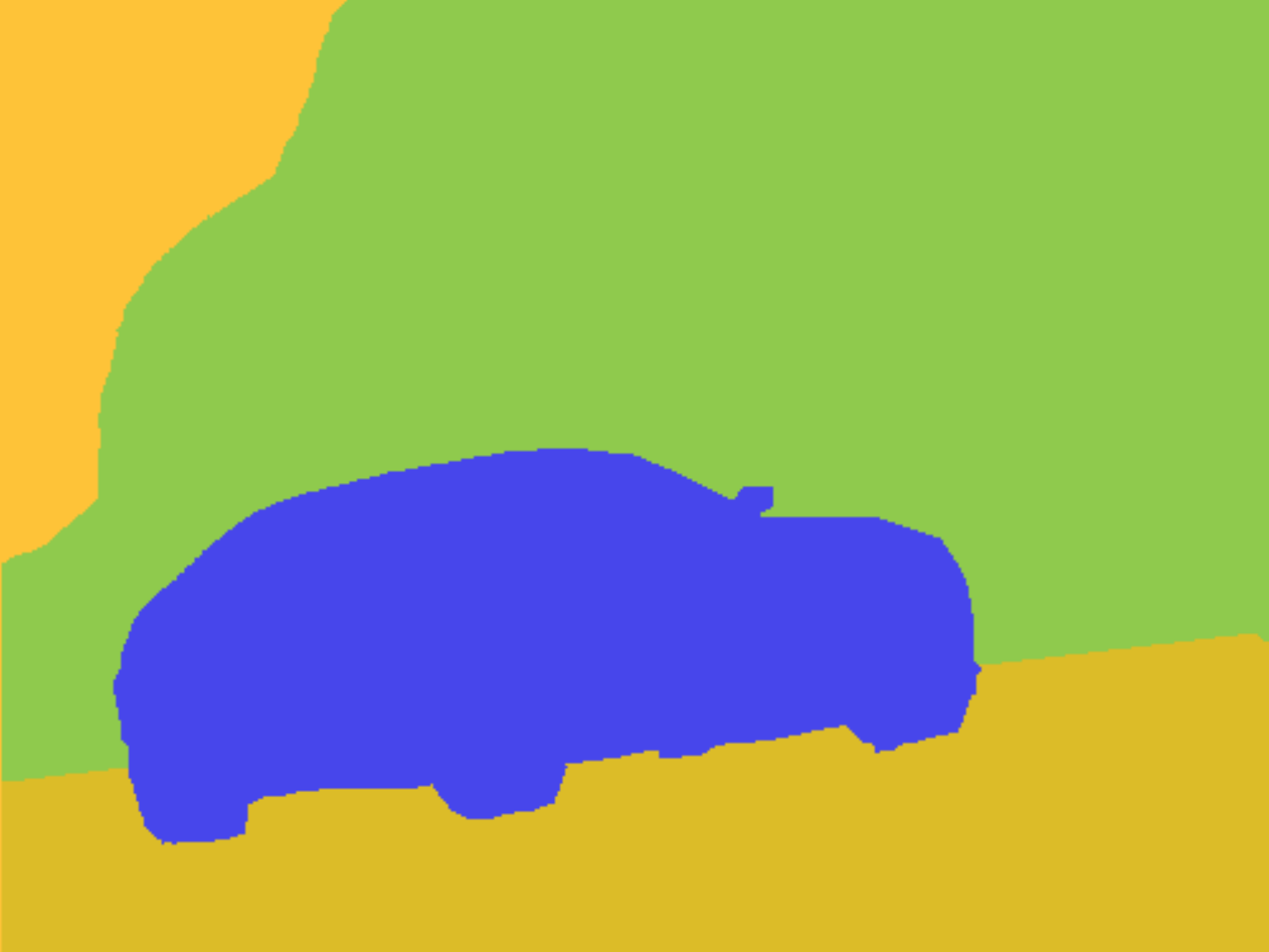}\\
(a) Image & (b) Truth\\
\includegraphics[width=1.6in]{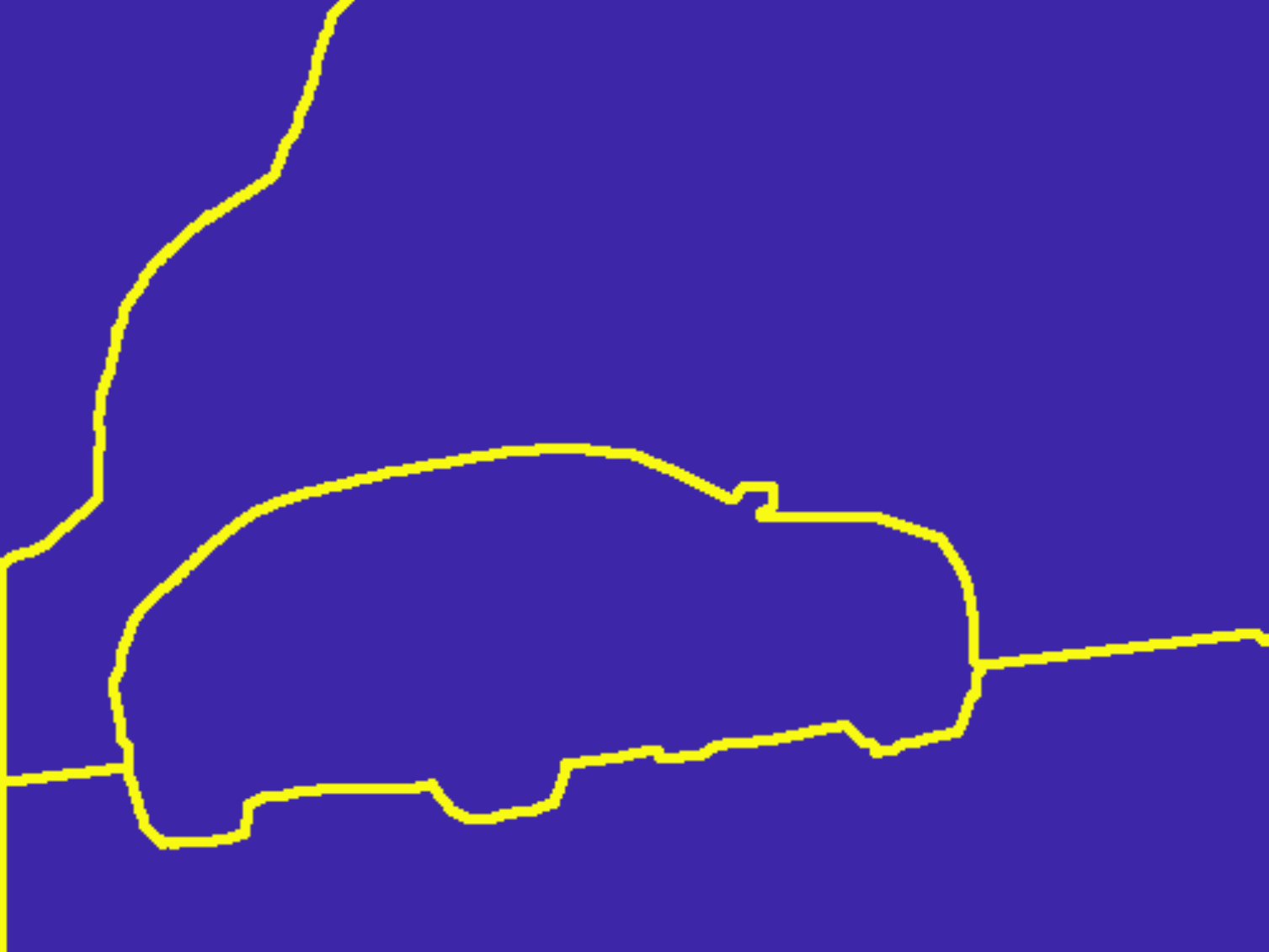}&
\includegraphics[width=1.6in]{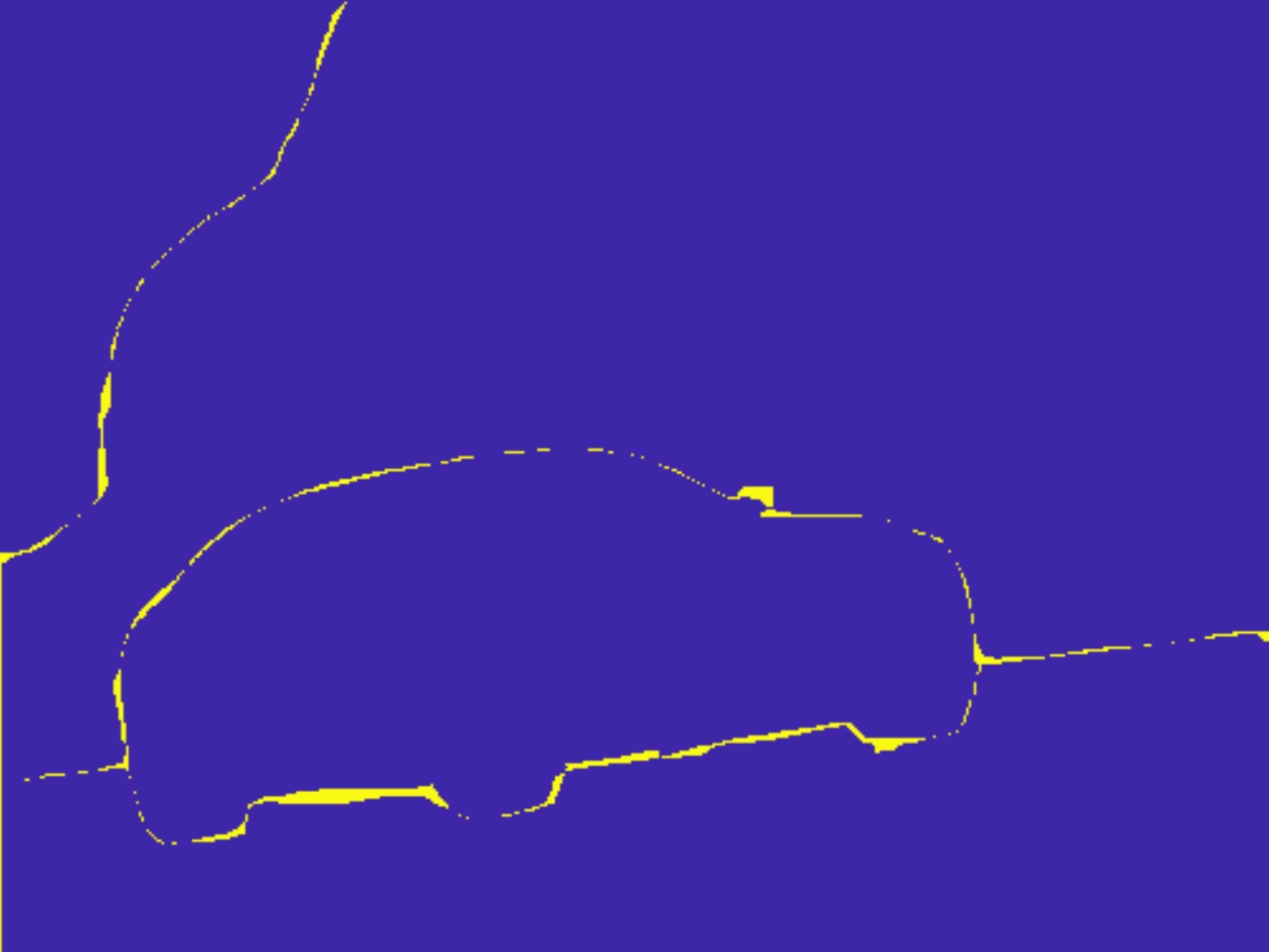}\\
(c) Truth borders & (d) Classifier errors\\
\end{tabular}
\end{center}
\caption{Region border classification errors. (a) Input image. (b) Ground truth regions. (c) Region borders. (d) Pixels mislabeled by the classifier.}
\label{fig:border-issues}
\end{figure}

{\bf Region Border Artifacts}. 
Region border labeling artifacts are common to all semantic segmentation classifiers due to imprecise manual ground truth labeling along the borders of objects, spatial pooling, etc.  
For example, in Fig.~\ref{fig:border-issues}(a) we show an example training image, and in Fig.~\ref{fig:border-issues}(b) we present its ground truth region labels. The ground truth region borders (i.e., edge map) are shown in Fig.~\ref{fig:border-issues}(c). The pixel labeling errors on this image made by the segmentation classifier RefineNet-Res101 \cite{Lin3} (trained on the dataset that includes this image) are highlighted in Fig.~\ref{fig:border-issues}(d). As is typical with other semantic segmentation classifiers, we consistently find labeling errors along the borders of regions. In this example, the classifier performed particularly well for the image as a whole, with nearly all of the errors occurring only at the region borders (other images will show additional larger errors {\em within} regions).

When learning the classifier/truth confusions, these border pixel errors should be ignored as they can pollute the desired statistics. They are merely artifacts that should not be considered. Therefore we limit the computation of the $P(C=c | l)$ confusions to only non-border pixel sites to avoid negatively influencing the confusion probabilities. 

To accomplish this, we first find all ground truth region border pixels in each image, then perform a spatial dilation of those border pixels. Only the complement of this pixel set in the image is employed for the computation of the confusion probabilities. The resultant confusion probabilities are  computed across all of the images to be analyzed and provides a single robust confusion representation for the entire dataset (not per image).

We additionally note that this underlying labeling issue at region borders therefore suggests that pixel-wise evaluations and comparisons of semantic segmentation approaches would be more appropriate if ignoring these pixels (which they currently do not discount). 
 

{\bf Training vs. Test/Validation Data.} It is also important to compute the confusion probabilities from data  {\em not} used to train the base classifier $C$. The confusions from training data will not be well representative of the types of confusions that will occur on additional (validation, test) data. 

To illustrate, we begin by representing the confusion probabilities $P(C=c | l)$ as an $|\mathcal{L}| \times |\mathcal{L}|$ matrix with the rows representing the class labels $c \in \mathcal{L}$ given by the base classifier output (best class hypothesis) and the columns signifying the ground truth classes $l \in \mathcal{L}$ (same set of classes). Therefore each $(m,n)$ cell in the matrix holds the value $P(C=m | n)$. In a scenario with absolutely no confusions, this matrix would be strictly diagonal. Any off-diagonal entries denote errors that exist in the output of the classifier. Note that this matrix is different from the earlier refinement matrix $R$.

\begin{figure}[t]
\begin{center}
\setlength\tabcolsep{.01 pt}
\begin{tabular}{ccc}
\includegraphics[width=1.75in]{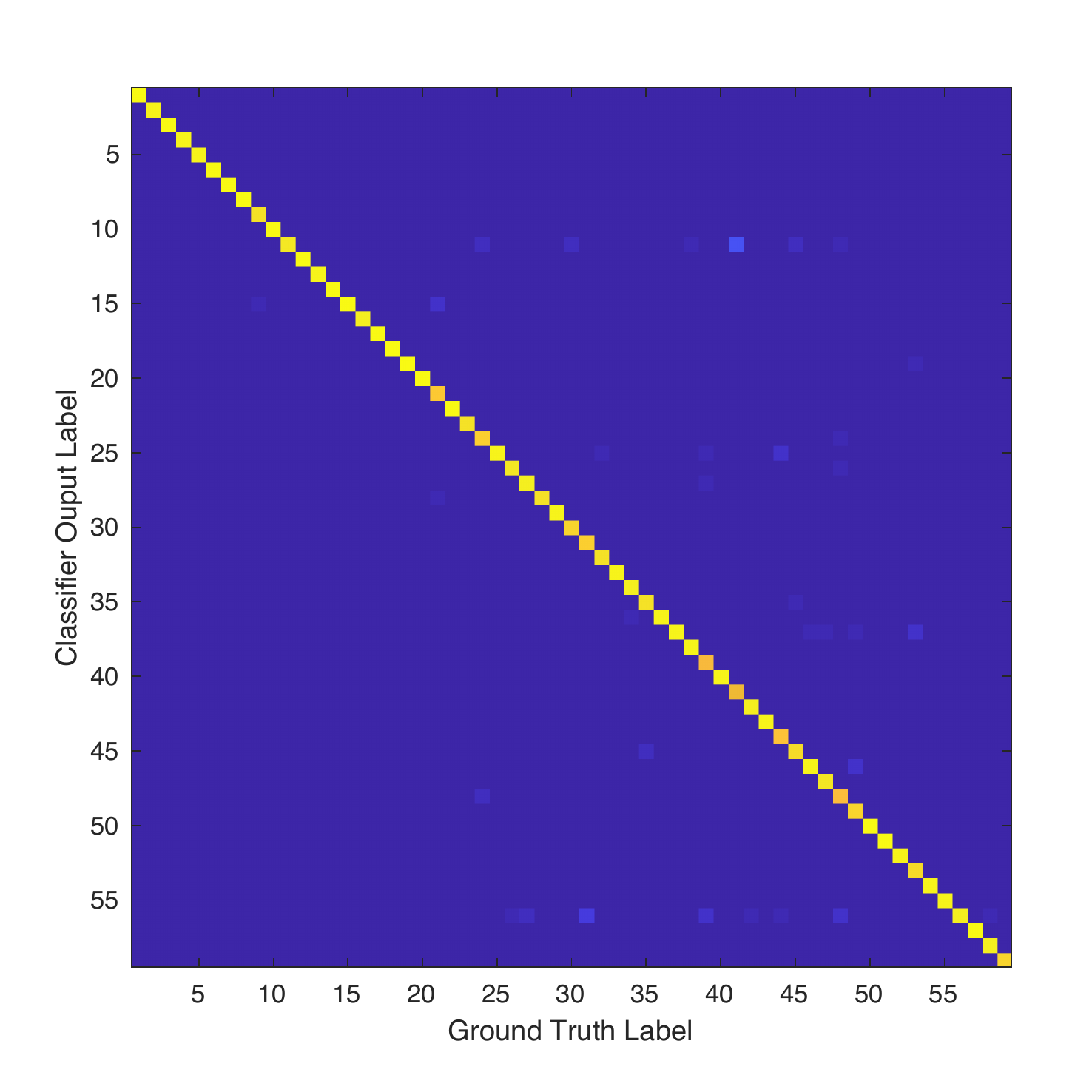}&
\includegraphics[width=1.75in]{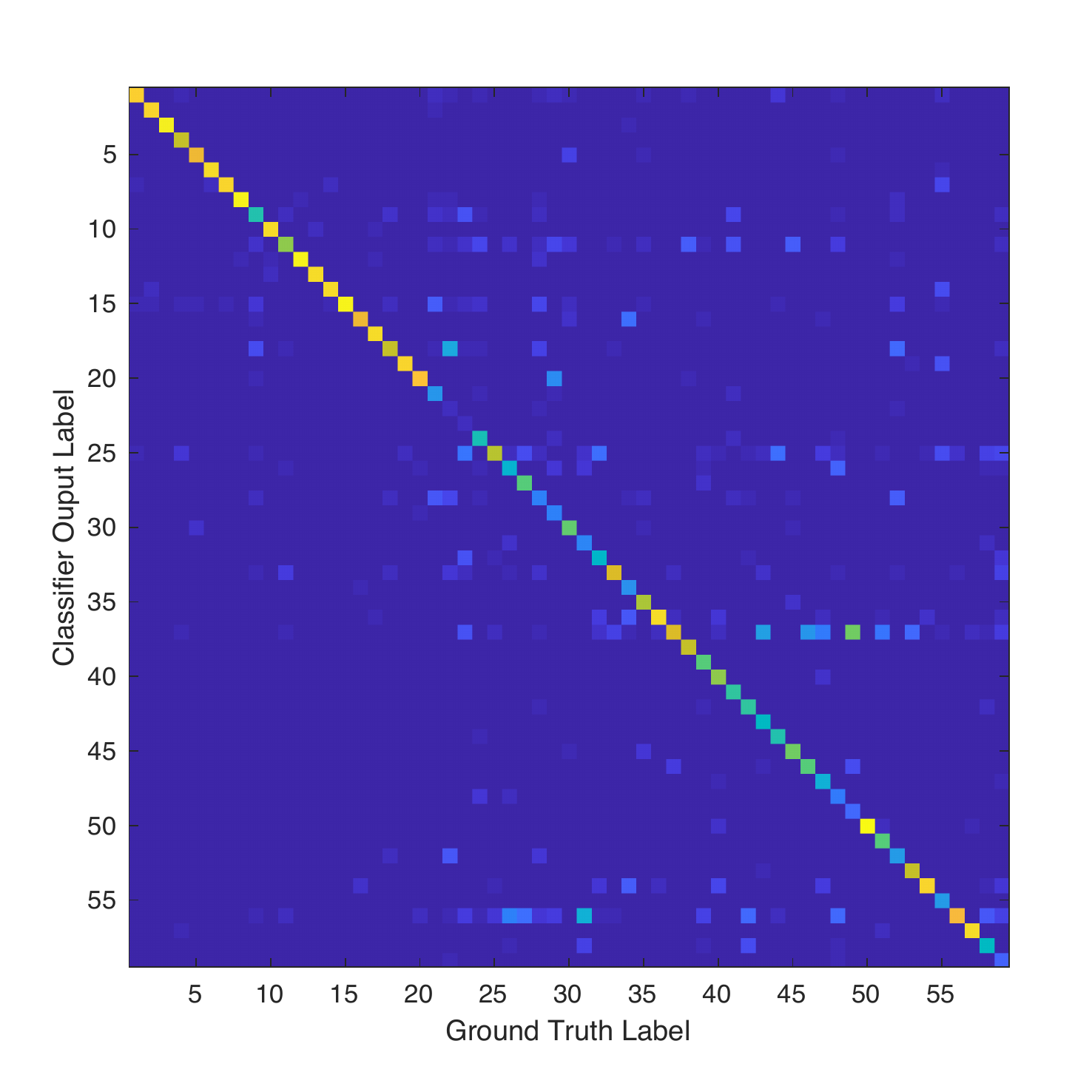}\\
(a) Training confusions & (b) Testing confusions \\
\end{tabular}
\end{center}
\caption{Confusion probability matrices constructed from PASCAL-Context using RefineNet-101 on (a) training and (b) testing data.}
\label{fig:train-test-confusions}
\end{figure}

To demonstrate the need for computing these confusion probabilities from separate validation data (instead of the training data itself), we show in Fig.~\ref{fig:train-test-confusions}(a) the confusion probability matrix generated from the PASCAL-Context {\em training} set ($\sim$5K images) using RefineNet-Res101 (trained on this data). Notice the strong diagonal structure, showing that the classifier did indeed fit the training data well. However, to compare, that same trained classifier was employed on the PASCAL-Context {\em test} set ($\sim$5K images, with ground truth available) and the resulting confusion probabilities are shown in Fig.~\ref{fig:train-test-confusions}(b). These results show much greater off-diagonal confusions, demonstrating that the classifier is prone to making more and qualitatively different types of mistakes on unseen data. Hence, these confusion probability differences demonstrate the need to learn the confusions from examples separate from the training data, ideally from a validation set if one is available 

However, we have found that many semantic segmentation approaches do not follow the standard machine learning practice of withholding a separate validation set out of the training data. They instead tend to use all available training data and validate directly on the test set (as ground truth for test data is typically available). As previously stated, it is recommended for the proposed approach to retain a small set of  validation data (not used to train the base classifier) to learn the confusion probabilities. If no validation data is available when using a pre-trained model, a small random sample of test data could be employed.

\subsection{Defining Prior Expectations}
\label{SubSect:Priors}

With the confusion probabilities now computed, the only remaining component necessary for classification refinement is the label prior $P(l)$. This term represents a prior expectation of a given class label $l$, and can be defined in several ways. 

{\bf Uniform Prior:} The simplest naive approach would be to employ a non-informative uniform prior. We refer to this version as $P_{uniform}(l)=1/|\mathcal{L}|$. With this prior all effort of refining the classifier probabilities would be placed onto the learned confusions.

{\bf Global Prior:} The prior could also be interpreted {\em globally} as the probability of observing class $l$ for any randomly selected pixel across the dataset (using ground truth information). This prior can be computed simply by L1-normalizing the histogram of class labels collected throughout the entire training dataset. We refer to this prior as $P_{global}(l)$. However, this formulation may bias the refinement to the most dominant class labels in the dataset.

{\bf Binary Prior:} Alternatively, we could specify the prior {\em locally}, estimating it for each image individually. A binary prior has a constant value for those classes present in the image and zero elsewhere (with the L1-norm equal to 1). We refer to this prior as $P_{binary}(l)$. It should be easier to estimate this image-level prior than  dense per-pixel label assignments (i.e., semantic segmentation). This prior has also been explored in LabelBank \cite{Hu1}.

{\bf Histogram Prior:}  Similar to the binary prior, we can construct a more informative non-binary prior using the L1-normalized histogram of classes present in the image. We refer to this prior as $P_{hist}(l)$. We posit that estimating this class-frequency histogram of labels for an image will be more difficult than for the binary prior, but similarly should be easier to estimate than the full per-pixel label assignments of an image.

{\bf Unconstrained Prior:} Lastly, the prior could be viewed more abstractly, separated from any binary or frequency-based constraint on its interpretation. In this case, for a given collection of annotated and classified sites in an image along with the confusion probabilities, the best possible prior could be solved using non-linear multi-variate techniques by minimizing the negative log-loss of the final refined class probabilities for the target ground truth labels. Based on  Eqn.~\ref{eqn:final-refine}, this is given by solving for $P(l)$ within the loss
\begin{equation}
Loss  =  -\sum_{i=1}^{\#sites} \log\left(\max\left[P(l_{i}^{gt} | d_i) , \epsilon\right]\right)
\end{equation}

\noindent where $\epsilon = 10^{-10}$. In our experiments with this prior, we used Matlab's {\tt \it fmincon} solver with the interior-point algorithm. We refer to the resulting (solved) unconstrained prior as $P_{uc}(l)$. 

We will examine and compare these different prior formulations in  Sect.~\ref{Sect:Experiments}, though we note that other definitions/interpretations of the priors may be possible.

\section{Experiments}
\label{Sect:Experiments}

We present experimental results  and analyses of the proposed refinement approach across 3 challenging datasets commonly used for semantic segmentation. ADE20K \cite{Zhou17} is a recent dataset of varied scenes with 20K training and 2000 testing images having a dense labeling of 150 classes. PASCAL-Context \cite{Mottaghi14} is a densely-labeled dataset of various scenes with 59 labels and is comprised of  4998 training and 5105 testing images. NYUDv2 \cite{Silberman1} contains RGB-D images (RGB only is analyzed) of indoor rooms (bedrooms, kitchens, bathrooms, etc.) with 795 training and 654 testing images (we employed the typical $425\times560$ image crop, as provided by \cite{Gupta1}). 

We selected the RefineNet \cite{Lin3} CNN semantic segmentation model as the base classifier to be refined with our approach. RefineNet is a ResNet-based architecture that employs residual chained pooling to localize coarse predictions using high-resolution features from multiple paths. This model has been previously trained on the aforementioned datasets and has been publicly-released online (RefineNet-Res101 trained on PASCAL-Context \& NYUDv2, and RefineNet-Res152 trained on ADE20K). Note that RefineNet includes the background (void) class as a valid output label, but in this work we removed background label from their outputs and renormalized the remaining label confidences.

\subsection{Confusion Probabilities}
We first present the computed confusion probability matrices for the three datasets. As discussed in Sect.~\ref{Sect:Approach}, the use of training (instead of validation) data to compute the confusion probabilities $P(C=c|l)$ can lead to poor performance. Since validation hold-out sets were not used by RefineNet for any of the datasets, we instead used the suggested random sample of test data (as pseudo-validation) to compute the confusion probabilities for each dataset. The number of randomly sampled test images for each dataset was 500 (of 2000) for ADE20K, 1000 (of 5105) for PASCAL-Context, and 200 (of 654) for NYUDv2. 

\begin{figure}[t]
\begin{center}
\setlength\tabcolsep{.01 pt}
\begin{tabular}{cc}
\includegraphics[height=1.6in]{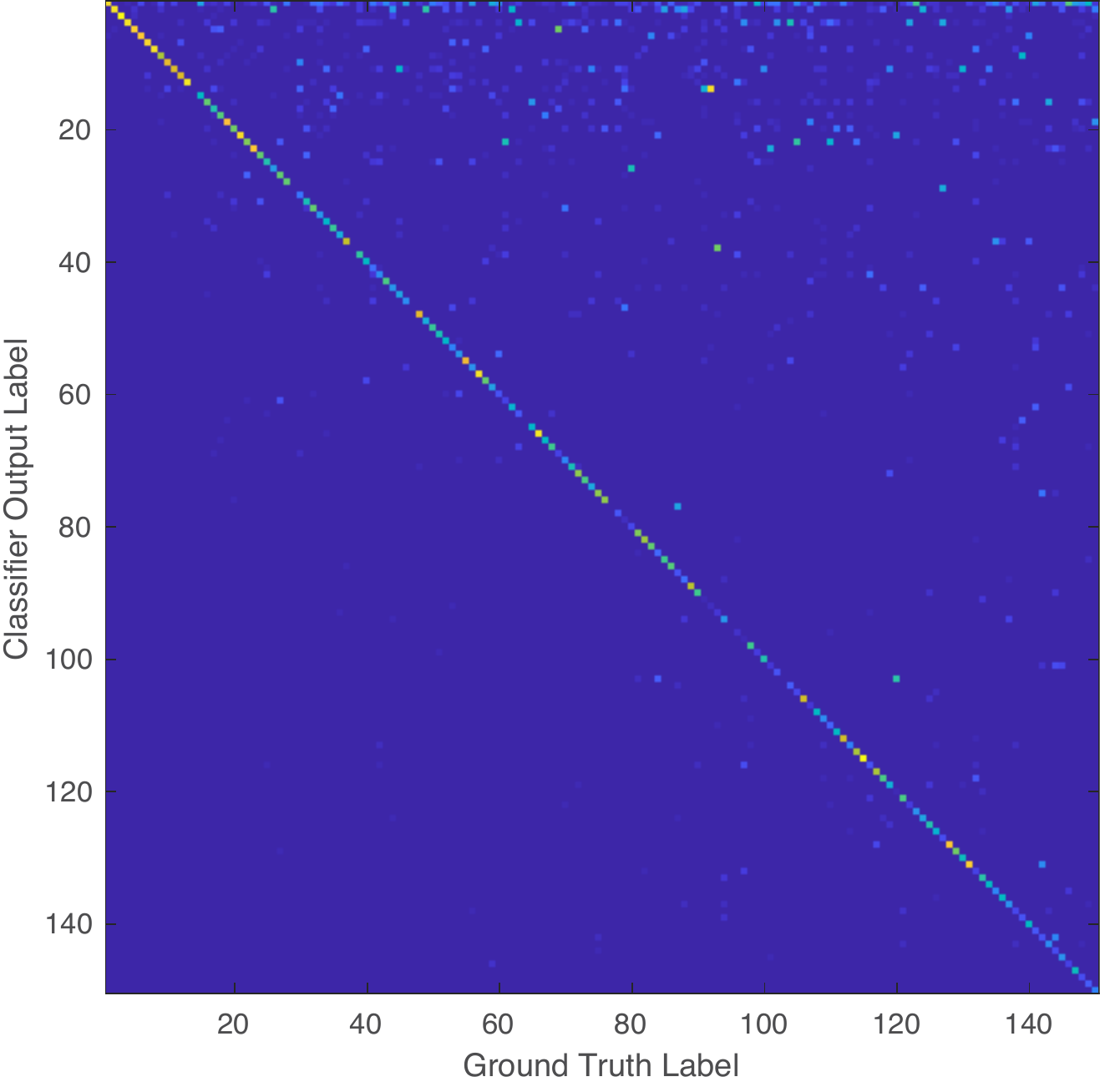} & 
\includegraphics[height=1.6in]{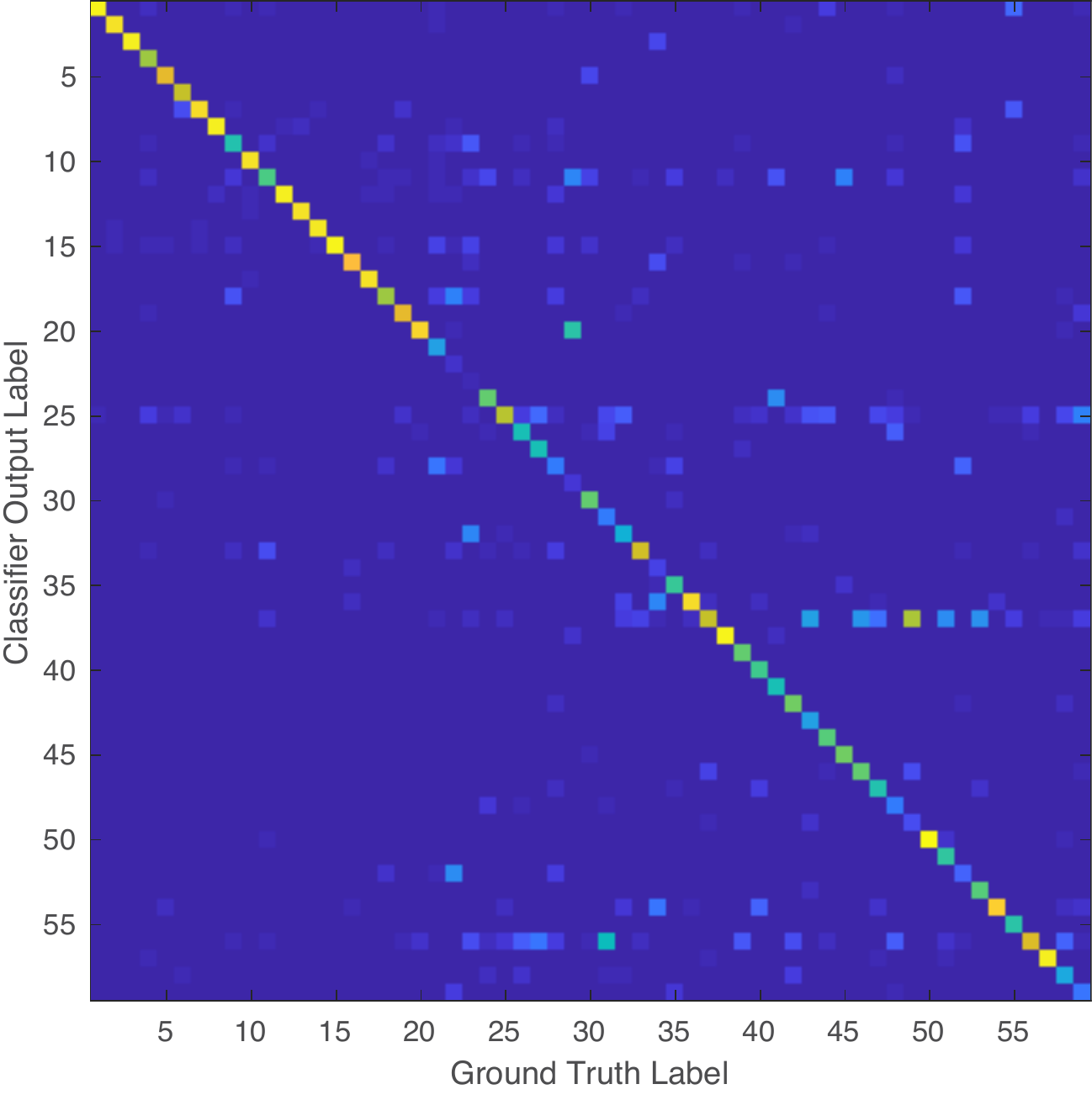}\\(a) ADE20K &
(b) PASCAL-Context \\
\multicolumn{2}{c}{\includegraphics[height=1.6in]{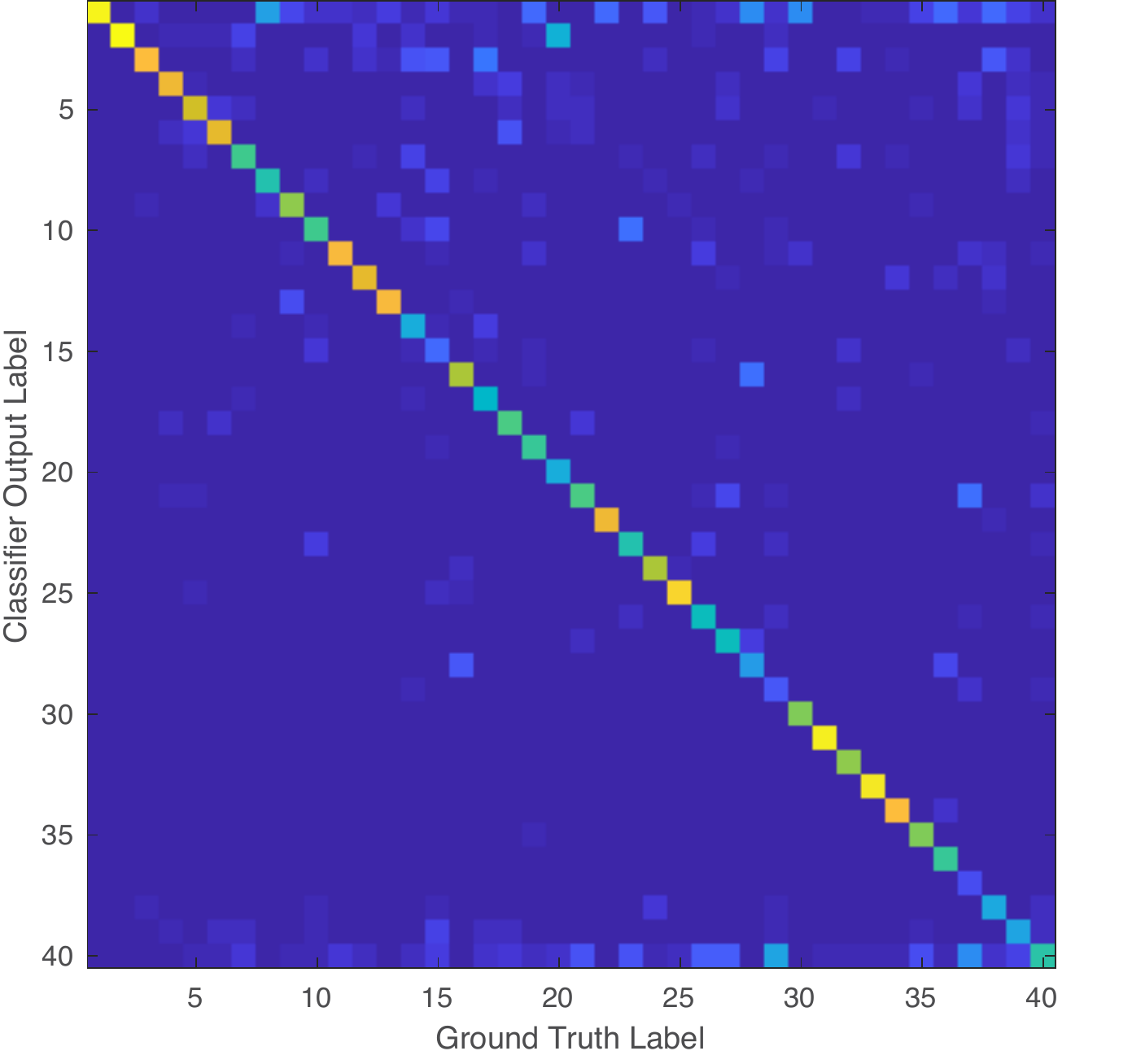}}\\
\multicolumn{2}{c}{(c) NYUDv2}\\
\end{tabular}
\end{center}
\caption{Confusion probability matrices generated using test data samples for the examined datasets.}
\label{fig:datasets-confusions}
\end{figure}

The resulting confusion probability matrices using the approach outlined in Sect.~\ref{SubSect:confusion-probabilities} are shown in Fig.~\ref{fig:datasets-confusions}. There are clearly many off-diagonal confusions present in each dataset. Some example confusion situations include \{{\tt animal}/{\tt rug}, {\tt river}/{\tt water}\} in ADE20K, \{{\tt sofa}/{\tt computer}, {\tt ground}/{\tt sidewalk}\} in PASCAL-Context, and \{{\tt door}/{\tt wall}, {\tt box}/{\tt other-prop}\} in NYUDv2. We also note the similarity of the sampled confusion probabilities for PASCAL-Context in Fig.~\ref{fig:datasets-confusions}(b) to the confusions derived from the entire test set (shown in Fig.~\ref{fig:train-test-confusions}(b)).

\subsection{Refinement with Different Priors}

We analyzed the standard score metrics of pixel accuracy and mean Intersection-over-Union (IoU) for various refinements of the base classifier using the different priors described in Sect.~\ref{SubSect:Priors}. The resulting scores from the base classifier on the three datasets are presented in the top row of Tables \ref{table:upper-pixacc} \& \ref{table:upper-iou}. 

Results using the non-informative uniform prior $P_{uniform}(l) = 1/|\mathcal{L}|$ and global prior $P_{global}(l)$ with the corresponding dataset confusion probability matrix (from Fig.~\ref{fig:datasets-confusions}) are provided in the second and third rows of the tables. The two methods are generally within $\pm 1\%$ of the base classifier, hence there is no significant change after refinement (as expected). The pixel accuracy scores for $P_{global}(l)$ are higher than $P_{uniform}(l)$, which is a result of the bias to more dominate classes in the dataset. The lower mean IoU scores for $P_{global}(l)$ also reflect this bias.

	\begin{table*}
		\begin{center}
		\begin{tabular}{|l| c | c | c|} 
			\hline
			 & ADE20K & PASCAL-Context & NYUDv2 \\ [0.5ex] 
			\hline
			Base Classifier (RefineNet \cite{Lin3}) & 78.6\% & 78.2\% & 72.9\% \\ 
            \hline
            Uniform Prior $P_{uniform}(l)$ & 77.0\% & 64.6\% & 72.7\% \\
            Global Prior $P_{global}(l)$ & 80.0\% & 77.7\% & 72.9\% \\ \hline
            LabelBank \cite{Hu1} (upper bound)  & 83.7\% & 85.7\% & 76.8\% \\ \hline
            Binary Priors $P_{binary}(l)$ (upper bound) & 85.9\% & 87.0\% & 78.1\% \\
			Histogram Priors $P_{hist}(l)$ (upper bound) & 88.1\% & 88.4\% & 80.3\% \\
            Unconstrained Priors $P_{uc}(l)$ (upper bound) & 89.0\% & 90.2\% & 82.6\% \\
			\hline 
		\end{tabular}
		\caption{Pixel accuracy scores on test data before/after various refinements.}
		\label{table:upper-pixacc}
        \end{center}
	\end{table*}

	\begin{table*}
		\begin{center}
		\begin{tabular}{|l| c | c | c|} 
			\hline
			 & ADE20K & PASCAL-Context & NYUDv2 \\ [0.5ex] 
			\hline
			Base Classifier (RefineNet \cite{Lin3}) & 34.0\%& 50.5\% & 45.2\% \\ 
            \hline
            Uniform Prior $P_{uniform}(l)$ & 35.1\% & 49.8\% & 45.5\% \\
            Global Prior $P_{global}(l)$ & 34.6\% & 49.1\% & 43.8\% \\ \hline
            LabelBank \cite{Hu1} (upper bound)  & 44.9\% & 62.0\% & 53.6\% \\ \hline
            Binary Priors $P_{binary}(l)$ (upper bound) & 53.6\% & 65.1\% & 55.9\% \\
			Histogram Priors $P_{hist}(l)$ (upper bound) & 57.7\% & 68.7\% & 59.2\% \\
            Unconstrained Priors $P_{uc}(l)$ (upper bound) & 59.3\% & 72.2\% & 62.5\% \\
			\hline
		\end{tabular}
		\caption{Mean IoU scores on test data before/after various refinements.}
		\label{table:upper-iou}
        \end{center}
	\end{table*}

We next examined the use of the more informative image-based priors, where we present the {\em ceiling} (best possible performance) metrics of refinement across the datasets. We used the ground truth data available to compute the optimal priors for each definition, leaving estimation of these particular priors to future work. This study is used to demonstrate the full potential of the proposed strategy and motivate future efforts.

For a relevant baseline of the image-based priors, we first show (in the fourth row of the tables) the upper bound ceiling results possible with the LabelBank \cite{Hu1} method. This approach removes (only) the out-of-context label errors using the optimal binary mask on the output label probabilities.  We see a significant improvement in the ceilings of both pixel accuracy (3.9-7.5\%) and mean IoU (8.4-11.5\%) over the base classifier. We note that this method is equivalent to our refinement approach using the optimal/known binary priors $P_{binary}(l)$ with an identity confusion probability matrix (no confusions).

Employing the binary priors $P_{binary}(l)$, as above, but now with the computed confusion matrix, performance is  improved. The fifth row of the table shows an increase of 1.3-2.2\%  for pixel accuracy and 2.3-8.7\% for mean IoU above the previous optimal LabelBank method (uses no confusion information). This shows that use of the confusion probabilities indeed can be used to increase the performance by additionally removing many {\em in}-context errors (not possible with the LabelBank masking method).

Using the histogram priors $P_{hist}(l)$ provides a further increase above the $P_{binary}(l)$ results, with an additional increase of
1.4-2.2\% for pixel accuracy and 1.6-3.6\% in mean IoU. This is expected, given the prior is better adjusted to reflect the true distribution of labels in each image. 

Lastly, the unconstrained priors $P_{uc}(l)$ give the expected best upper bound performance of all the prior versions, yielding an additional increase over $P_{hist}(l)$ of 0.9-2.3\% in pixel accuracy and 1.6-3.3\% in mean IoU. As these priors are solved to optimize the given refinement formulation, the improvements should be close to the maximum theoretically possible results.

\subsection{Discussion and Future Directions}

Overall, we see possible refinement gains over the base classifier of 
9.7-12.0\% in pixel accuracy and 17.3-25.3\% for mean IoU across the datasets. These upper bound results show that there is still much room for improvement in the classifier that can be achieved using refinement with the confusion probabilities and appropriate priors. When the prior is defined {\em globally} (independent of the image), as in $P_{uniform}(l)$ or $P_{global}(l)$, and used together with the confusion probabilities to refine the test images, there is minimal benefit (or even a slight loss in some datasets). The notable gains are achieved when the priors are computed independently for each test image. The presented results with the image-based priors $P_{binary}(l)$, $P_{hist}(l)$, and $P_{uc}(l)$ demonstrate that increasingly higher gains are possible in both pixel accuracy and mean IoU with the proposed strategy.

Based on these results, future work will address in more detail how particular prior expectations can be reliably estimated from image data. One may consider constructing the priors from the base classifier's existing semantic segmentation results (e.g., determine $P_{binary}(l)$ or $P_{hist}(l)$ from the histogram of labels given by the classifier for a test image), but this refinement yields no improvement as it is obviously biased to the base classifier results. One possible method may be to find other similar training images for a test image and use a weighted combination of their priors, as images that appear similar likely have related label priors. Another  path to explore is a specialized deep learning technique to estimate the prior directly from the input image itself. Unlike the semantic segmentation task (or even image classification) which is typically a trained to 1-hot categorical vectors of class labels, the target outputs here are multinomial distributions. 

Though an approach already exists in \cite{Hu1} to estimate $P_{binary}(l)$, the normalized label histogram prior $P_{hist}(l)$ shows a higher performance ceiling and hence a method to estimate this prior would be more beneficial. We expect the unconstrained priors $P_{uc}(l)$, which resulted in the highest performance metrics, will be more difficult to estimate from test images as the priors are a function of both the image labels present and the base classifier confusion probabilities. Hence we suggest that estimation of the histogram-based priors $P_{hist}(l)$ may be the best focus for future research.

As the proposed method continues to produce label probabilities/confidences, it can therefore be easily integrated with other processing strategies or further context-based reasoning. For example, instead of probabilistically refining each pixel site independently of other sites, a more holistic approach, such as formulating a spatial random field (e.g., CRF), could be employed to produce a more spatially-consistent solution.

\section{Conclusion}
\label{Sect:Conclusion}

Semantic segmentation is a complex visual task requiring reliable and consistent pixel-wise label predictions. Current models produce systematic errors which can be exploited to refine the initial confidences of the classifier. We presented a general Bayesian strategy for classifier refinement, deriving a straight-forward and compact description comprised of confusion probabilities, label priors, and classifier confidences. By understanding the types of mistakes made in the classifier label assignments, we demonstrated the strong potential of the proposed method by successfully refining a state-of-the-art classification model on multiple datasets. The results further motivate future research into reliably estimating prior expectations from images.

\section*{Acknowledgment}

This research was supported by the United States Air
Force Research Laboratories under contract No. GRT00044839/60056946, Subcontract 60056946.

\bibliographystyle{IEEEtran}
\bibliography{mybib}

\end{document}